\documentclass[german,english]{bvm2019}

\title{Fully-deformable 3D image registration in two seconds}

\author{Daniel Budelmann$^1$, Lars K\"onig$^1$, Nils Papenberg$^1$, Jan Lellmann$^2$}

\authorrunning{Budelmann et al.}
\institute{%
$^1$Fraunhofer Institute for Medical Image Computing (MEVIS), L\"ubeck\\
$^2$Institute of Mathematics and Image Computing, Universit\"at zu L\"ubeck
}

\email{daniel.budelmann@mevis.fraunhofer.de}

\begin{document}

%
\selectlanguage{english}

\maketitle

\begin{abstract}
We present a highly parallel method for accurate and efficient variational deformable 3D image registration on a consumer-grade graphics processing unit (GPU). We build on recent matrix-free variational approaches and specialize the concepts to the massively-parallel manycore architecture provided by the GPU. Compared to a parallel and optimized CPU implementation, this allows us to achieve an average speedup of $32.53$ on $986$ real-world CT thorax-abdomen follow-up scans. At a resolution of approximately $256^3$ voxels, the average runtime is $1.99$ seconds for the full registration. On the publicly available DIR-lab benchmark, our method ranks third with respect to average landmark error at an average runtime of $0.32$ seconds.

\end{abstract}

\section{Introduction}
Image registration -- i.e., finding a dense correspondence map between images or volumes taken at different points in time or under different conditions -- is still a crucial component of many clinical and research pipelines: compensating for patient movement and breathing in radiological follow-up and radiation therapy, monitoring progression of degenerative diseases, 3D reconstruction from slices in histopathology, and many others. It is made particularly challenging by the typically large, three-dimensional nature of the data, highly non-convex energies, and runtime requirements of clinical practice.


Towards reducing runtime, the authors of \cite{2653-01,2653-02} propose a highly accurate non-rigid registration model with applications in follow-up imaging in radiology and liver ultrasound tracking, and introduce a parallel algorithm for the CPU. They also include a preliminary GPU implementation for the 2D case provided by \cite{2653-03}. In this work, we extend these ideas to a fast, matrix-free, parallel algorithm  that solves the variational, regularized problem for full 3D image registration on the GPU. This allows us to achieve sub-second runtimes on the standard resolution of $128^3$, and in the low seconds range for high-resolution $256^3$ volumes, at state-of-the-art accuracy.


%

\section{Materials and methods}
\subsection{Model}
\label{2653sub:regFramework}
Regarding the model, we follow \cite{2653-02}: We seek a three-dimensional deformation vector field $y\in\mathbb{R}^{3\overline{m}^y}$, $\overline{m}^y:=m_x^y m_y^y m_z^y$, discretized on a\emph{ deformation grid} with dimensions $m_x^y\!\times\!m_y^y\!\times\!m_z^y$, which deforms a template image $\mathcal{T}\in\mathbb{R}^{\overline{m}}$ to be similar to a reference image $\mathcal{R}\in\mathbb{R}^{\overline{m}}$, $\overline{m}:=m_x m_y m_z$, both discretized on an \emph{image grid} with dimensions $m_x\!\times\!m_y\!\times\!m_z$.

To find $y$, we numerically minimize an \emph{objective function} $\mathcal{J}(y)~:~\mathbb{R}^{3\overline{m}^y}~\rightarrow~\mathbb{R}$, consisting of distance measure $\mathcal{D}$ and smoothing term $\mathcal{S}$, weighted by $\alpha>0$:
\begin{align}
\label{2653eq:joint}
y^{\ast} := \arg\min_{y\in\mathbb{R}^{3\overline{m}^y}} \mathcal{J}(y), \quad \mathcal{J}(y) := \mathcal{D}(\mathcal{R},\mathcal{T}(P(y))) + \alpha \mathcal{S}(y).
\end{align}
Here $P~:\mathbb{R}^{3\overline{m}^y}\rightarrow\mathbb{R}^{3\overline{m}}$ denotes the grid conversion $Py =: \hat{y}$, which converts the deformation $y$ from the deformation grid to the image grid, before it is used to interpolate the deformed template image $\mathcal{T}(P(y))$ on the image grid.

For the distance measure, we use the well-known \textit{normalized gradient field} (NGF), which is particularly suitable for multi-modal images \cite{2653-04}. It focuses on intensity changes and compares the angles of the image gradients. 
To encourage smooth deformations, we employ the curvature-based regularization term $\mathcal{S}(y)$ introduced by \cite{2653-05}, which penalizes the Laplacian of the deformation field components $y_i$ via $(\Delta y_i)^2$.

For solving \eqref{2653eq:joint} numerically and robustly without accurate initialization, we use the \textit{Limited-Memory Broyden-Fletcher-Goldfarb-Shanno} algorithm described in \cite{2653-06}, embedded in a multi-level (coarse-to-fine) approach.



\subsection{Parallelization}
We chose to implement our method on the GPU using the CUDA toolkit, which allows working close to the hardware and fine-tuning.


Performance on the GPU\ is tightly coupled to a high \textit{occupancy}, defined as the number of running threads divided by the number of potentially running threads that the device can handle. Using a large number of registers per thread decreases occupancy \cite{2653-07}, therefore, we keep the number of variables per thread low and split large functions (kernels) into smaller ones. 

We generally used single-precision ($32\ts$bit) floating variables due to the faster computations and only half the number of required registers compared to double precision ($64\ts$bit)~\cite{2653-07}. 

\paragraph{Generating the multi-level pyramid.}
For the multi-level approach, reference and template images need to be downsampled to various resolutions. The CUDA framework provides CUDA \textit{streams}, which enable concurrency between GPU computations and memory transfers between host and GPU \cite{2653-07}.
This allows to run the pyramid generation and data transfer for reference and template image in parallel.

\paragraph{Evaluating the objective function.}
\label{2653sub:minimizing}
 Evaluating the distance term \mbox{$\mathcal{D}(y):~\mathbb{R}^{3\overline{m}^y}\rightarrow\mathbb{R}$} and its gradient requires two grid conversions and a gradient computation:
\begin{enumerate}
        \item convert the deformation $y$ to the image grid, denoted by $P$: $\hat{y}~:=~P(y)$,
        \item compute the distance measure $\mathcal{D}$ and its gradient $\nabla \mathcal{D}(\hat{y})$, and
        \item convert $\hat{y}$ and $\nabla \mathcal{D}(\hat{y})$ to the deformation grid by applying $P^\top$. 
\end{enumerate}
In the following sections, we discuss the details of each step and its implications for the implementation with CUDA.

\paragraph{Distance measure and gradient computation.}
\label{2653sub:ngf}
We denote by $\nabla \mathcal{R}_i$ and $\nabla \mathcal{T}_i(P(y))$ the gradients of the reference and deformed template image at the $i$-th image grid point and discretize the NGF distance measure as a sum over grid points,
\begin{equation}
\mathcal{D}_{\text{NGF}}(y)=\frac{\overline{h}}{2} \sum_{i=1}^{\overline{m}} \left(1- \left( \frac{\langle\nabla \mathcal{T}_i(P(y)), \nabla \mathcal{R}_i\rangle+\tau \varrho}{||\nabla \mathcal{T}_i(P(y))||_\tau ||\nabla \mathcal{R}_i ||_\varrho} \right)^2 \right), 
\end{equation}
with voxel volume $\overline{h} = h_x h_y h_z$ as product of the image grid spacings, the smoothed norm function $||\cdot||_\varepsilon=\sqrt{\langle\cdot,\cdot\rangle+\varepsilon^2}$, and the modality-dependent parameters $\tau, \varrho > 0$ to filter the gradient image for noise. 
Following \cite{2653-01}, we can parallelize the computation of the distance measure function value directly over the terms in the sum.

Applying derivative-based numerical optimization methods such as L-BFGS (section~\ref{2653sub:minimizing}) requires frequent evaluation of the gradient $\nabla \mathcal{D}$. The chain rule yields $\nabla \mathcal{D}_{\text{NGF}}(y) = \frac{\partial \psi}{\partial \mathcal{T}} \frac{\partial \mathcal{T}}{\partial P} \frac{\partial P}{\partial y}$ with the reduction function $\psi:\mathbb{R}^{\overline{m}}\rightarrow\mathbb{R}$. 

Evaluating the gradient using the chain rule by computing the gradient parts and multiplying step-by-step is expensive in terms of (intermediate) memory required. 
We avoid this by relying on the matrix-free methods introduced by \cite{2653-02}. 

\paragraph{Grid conversion.}
\label{2653sub:gridConv}
Following the approach proposed by \cite{2653-01}, we separate the deformation grid resolution $\overline{m}^y$ and the image resolution $\overline{m}$. This allows to save memory and speed up the registration by discretizing the deformation on a coarser grid while preserving all information in the input images.

For optimal performance, a (surprisingly) crucial step in computing the distance measure and its gradient is conversion between the two grids, i.e., computing matrix-vector products with $P$ and $P^\top$. Applying $P$ is directly parallelizable when using trilinear interpolation~\cite{2653-01}. However, applying $P^\top$ with a coarser deformation grid produces possible write conflicts introduced when summing up values from multiple points on the higher-resolution image grid in order to obtain a value for a single point on the lower-resolution deformation grid.

To account for this issue, the authors of \cite{2653-01} introduced a red-black scheme, where all odd slices are computed in parallel, followed by all even slices. However, the author of \cite{2653-03} observed a poor utilization of GPU cores with this method. Therefore they computed every slice, row, and column in parallel, and used atomic operations to avoid write conflicts.

We introduce a different method, which is not based on the red-black-scheme and free of write conflicts: Each thread computes a deformation grid point independently by summing the corresponding image domain points, 
\begin{align}
y = 
\sum_{i\in\Omega_1} 
\sum_{j\in\Omega_2} 
\sum_{k\in\Omega_3} \omega \cdot \hat{y}_{i,j,k}.
\end{align}
Here, $\omega$ is the local weight and $\Omega_i$ are the corresponding indices of $\hat{y}$ for each dimension, which are determined beforehand, separately for each dimension.
While there is a certain overhead in computing the weights $\omega$ this way, in our case it was found that the overall runtime is still faster due to the higher parallelism.
\section{Results}
\label{2653sub:evaluation}
We investigated the accuracy and speed of our method in comparison to state-of-the-art alternatives from the DIR-Lab 4DCT benchmark~\cite{2653-08,2653-09}. We also compared to an Open Multi-Processing--(OMP--)based implementation of the same model on the CPU proposed in \cite{2653-02}, which is already one to two orders of magnitude faster than a matrix-based implementation using the MATLAB FAIR toolbox~\cite{2653-04}.

All experiments were performed using an NVIDIA GeForce GTX 1080Ti GPU and an Intel Core i7-6700K CPU.





\subsection{Radboud follow-up CT dataset}
\label{2653sub:runtime}
In order to investigate the performance of our method on high-resolution 3D data, we measured the average runtime over 986 registrations on a dataset of follow-up thorax abdomen CT scans provided by the Radboud University Medical Center, Nijmegen, Netherlands. 
The images have resolutions in the range of $512^2 \times \{72, \ldots, 1577\}$. As full image resolution was slightly out of reach due to memory restrictions of the GPU, we evaluated our approach on half, quarter, eighth and sixteenth resolution per dimension.

For the highest resolution, average runtime was $1.99$ seconds, with an average speedup of $32.53$ compared to the CPU-based parallel OMP implementation (Table~\ref{2653tab:runtimes}). On the lower resolutions, our method achieves sub-second runtimes at a speedup of about one order of magnitude. A majority of the runtime on the lower resolutions is spent on the multi-level creation, due to the large memory transfer and downsampling.

\begin{table}
        \caption{Mean runtimes and standard deviations, averaged over 986 thorax-abdomen registrations. Finest image resolutions were approximately $256^3$, $128^3$, $64^3$ and $32^3$.
        Compared to the CPU-based OMP implementation, we achieve an average speedup of $32.53$ with average runtimes of less than $2$ seconds, which opens up new application scenarios for clinical use and interactive registration.
}
        \label{2653tab:runtimes}
        \centering
        \begin{tabular}{ccccc}
         \hline
         & $256^3$ & $128^3$ & $64^3$ & $32^3$ \\
         \hline
         Ours (s)& $\mathbf{1.99\pm0.87}$ & $0.56\pm0.14$ & $0.39\pm0.08$ & $0.36\pm0.08$ \\
         OMP (s) & $66.94 \pm 39.36$ & $ 8.11 \pm 3.21 $ & $2.24\pm0.69$ & $1.62\pm0.43$ \\
         Speedup & $\mathbf{32.53\pm10.04}$ & $14.06\pm2.56$ & $5.64\pm0.68$ & $4.51\pm0.40$ \\
         \hline
        \end{tabular}
\end{table}



\begin{figure}[b] 
        \centering
        \begin{subfigure}[]
                {\includegraphics[height=3.19cm]{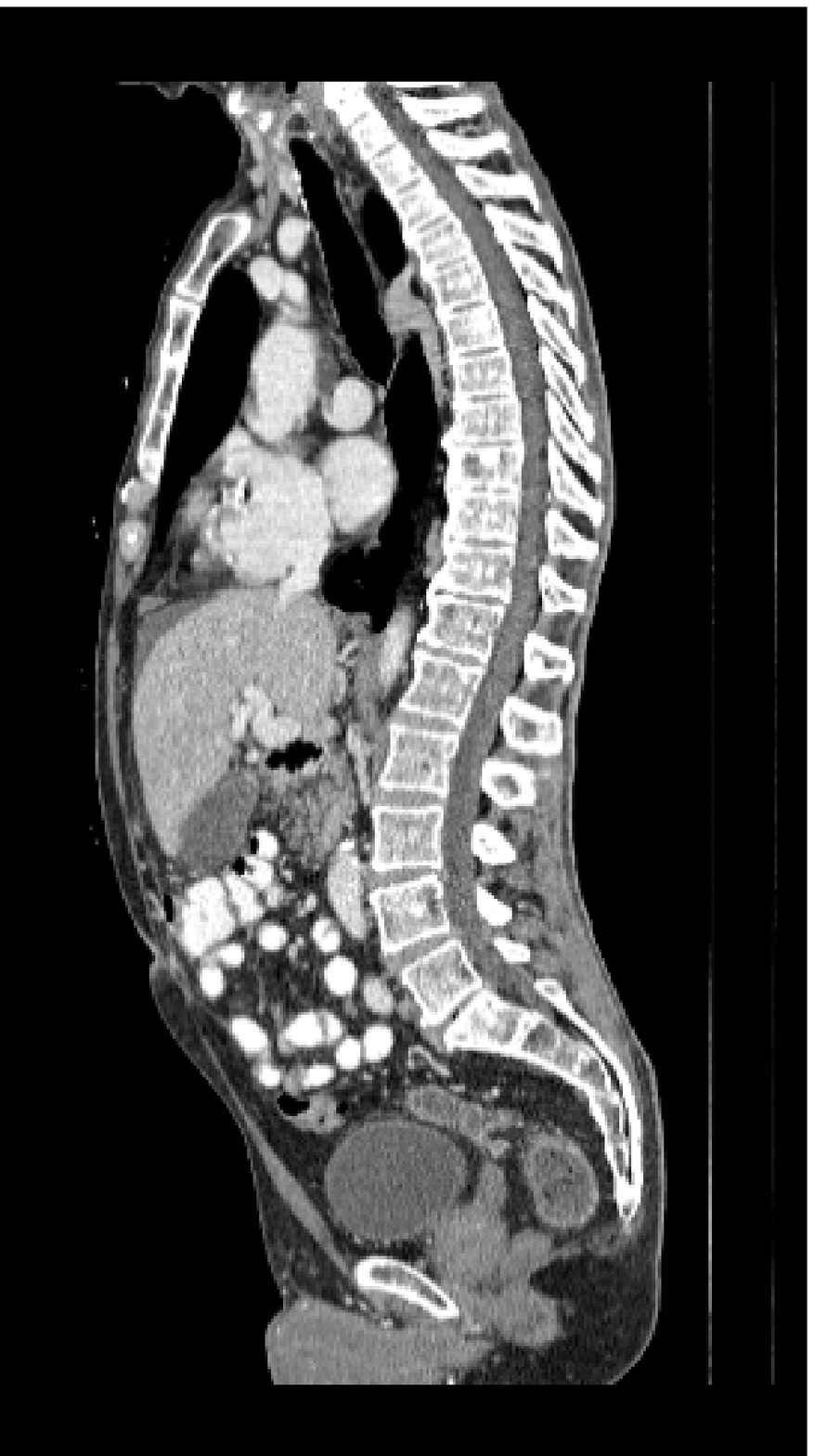}}
        \end{subfigure}%
        \begin{subfigure}[]
                {\includegraphics[height=3.19cm]{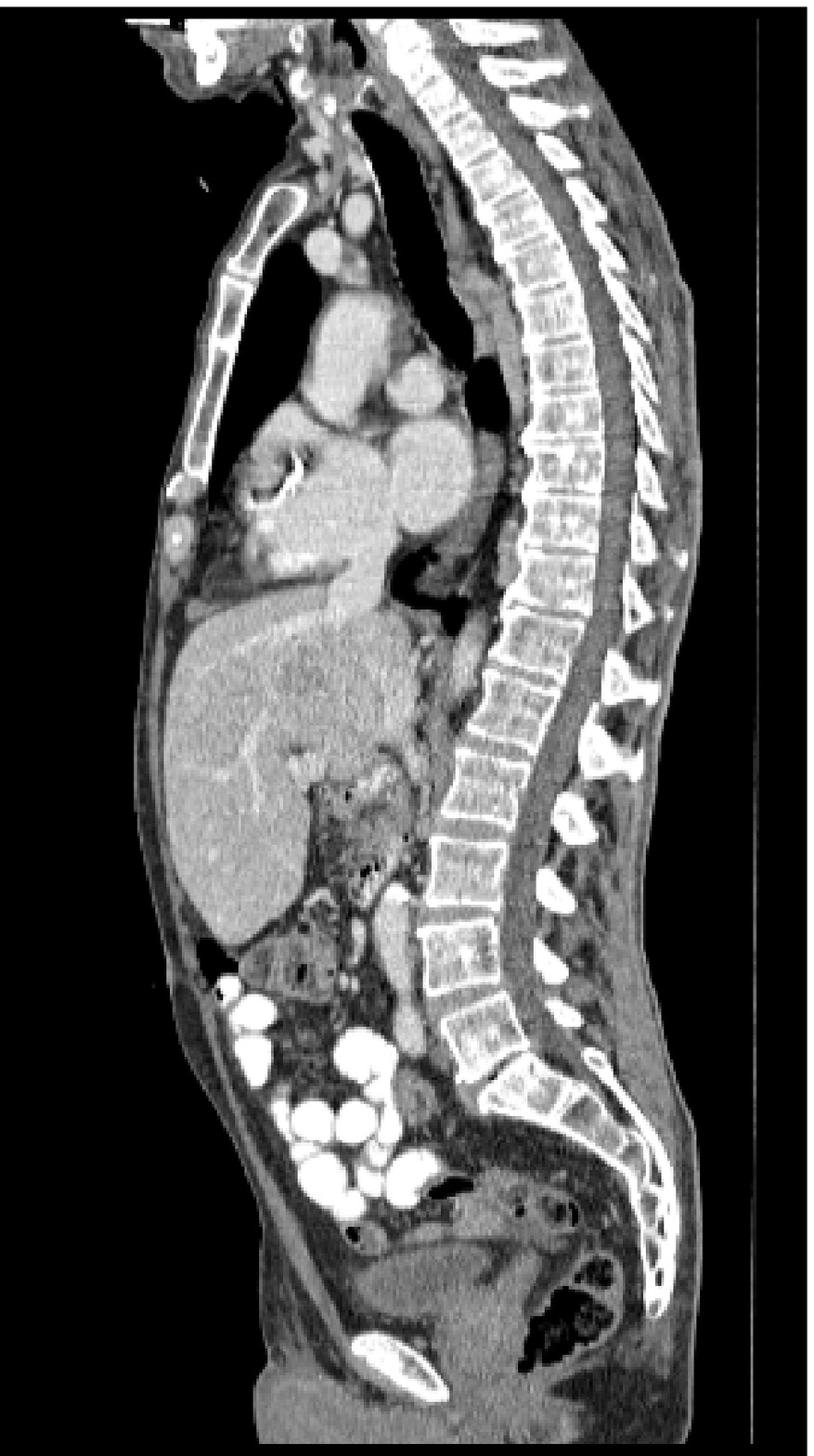}}
        \end{subfigure}%
        \begin{subfigure}[]
                {\includegraphics[height=3.19cm]{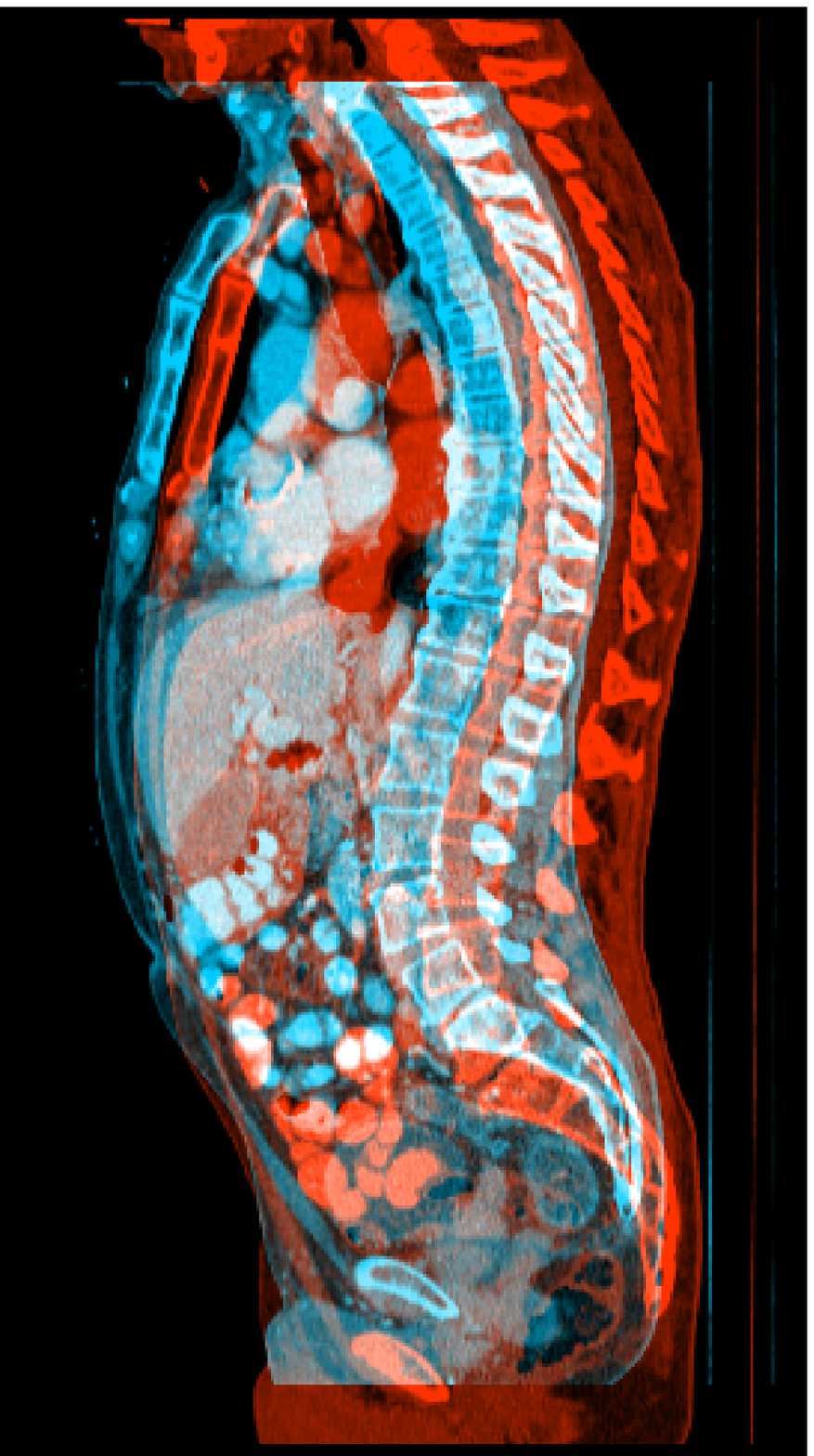}}
        \end{subfigure}%
        \begin{subfigure}[]
                {\includegraphics[height=3.19cm]{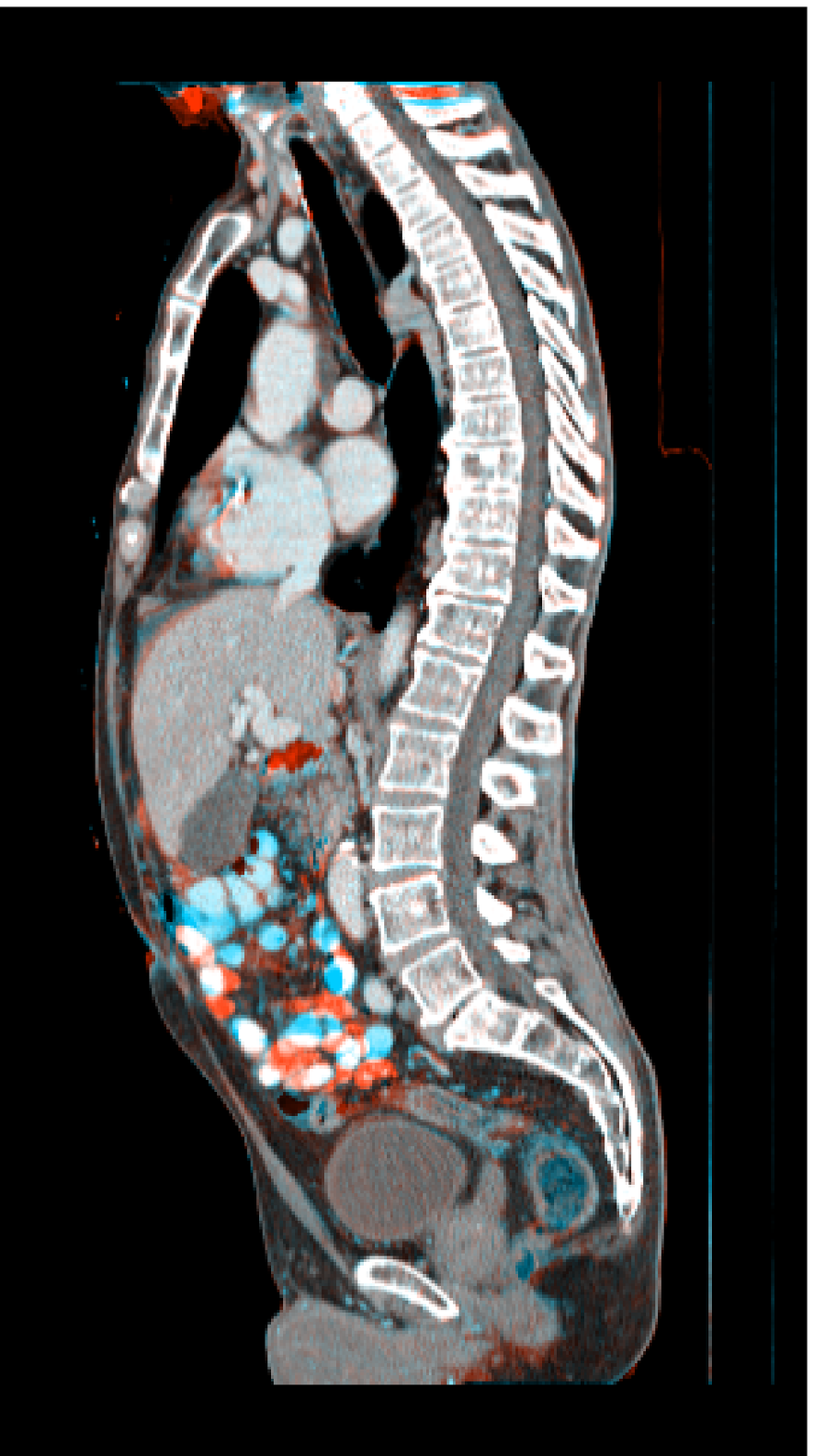}}
        \end{subfigure}%
        \begin{subfigure}[]
                {\includegraphics[height=3.19cm]{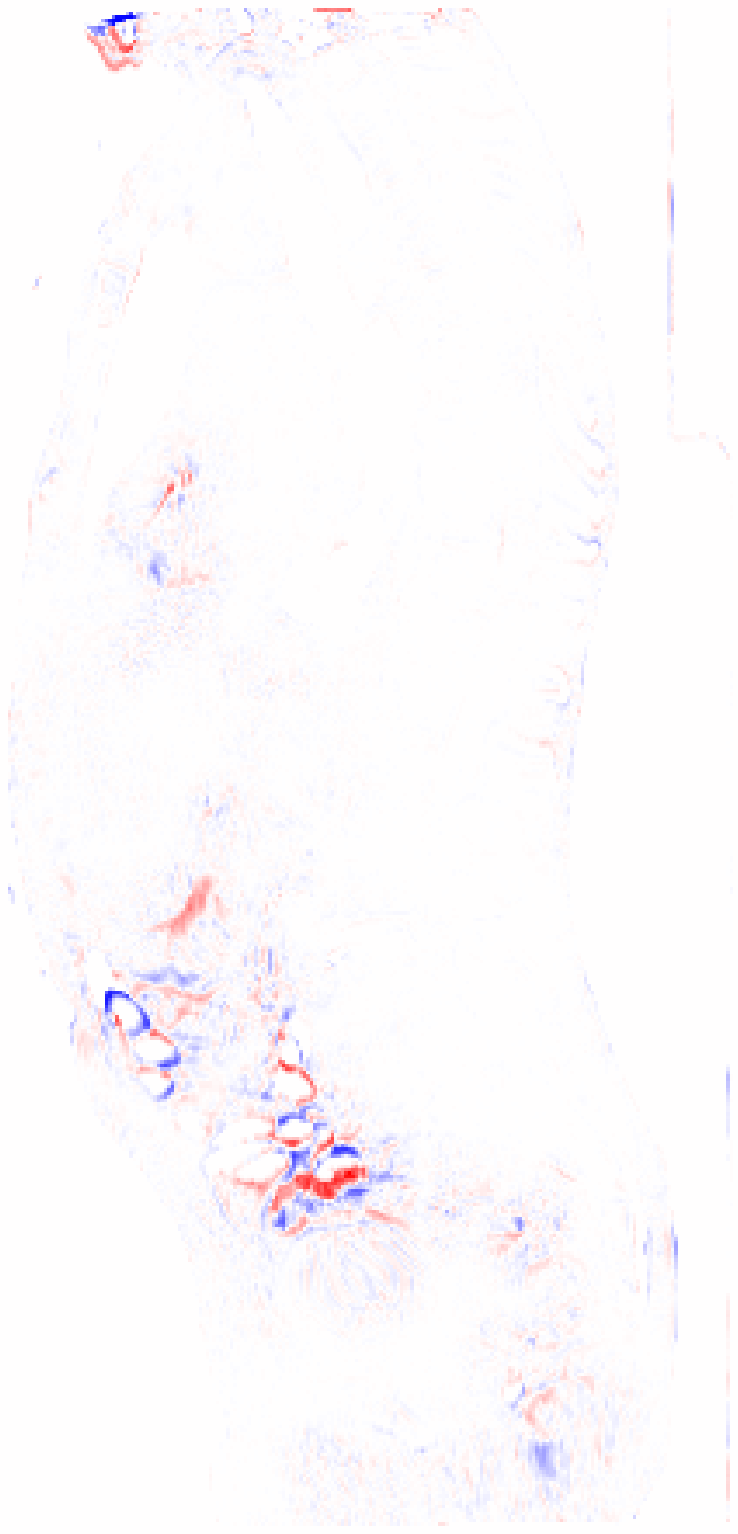}}
        \end{subfigure}%
        \begin{subfigure}[]
                {\includegraphics[height=3.19cm]{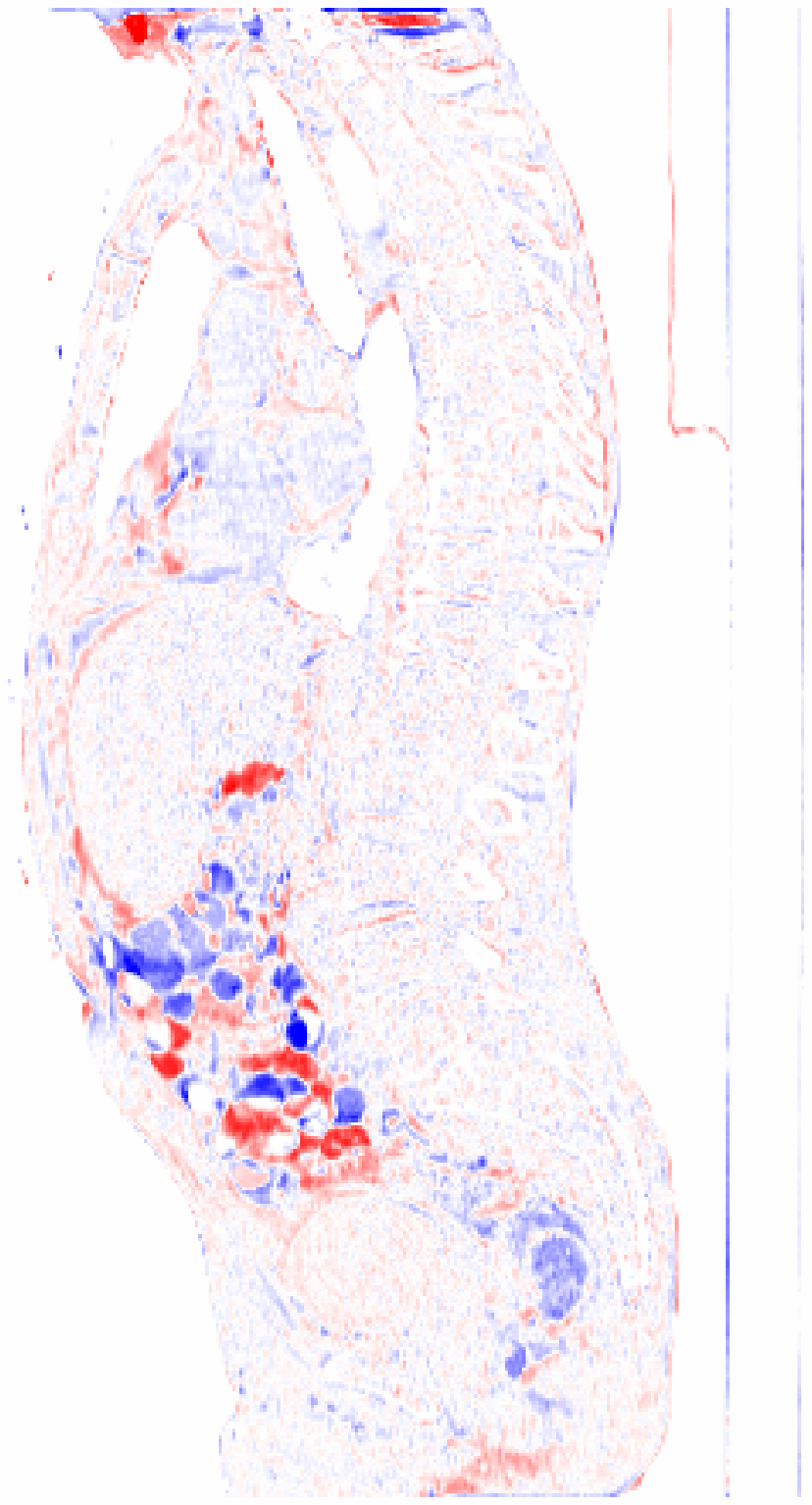}}
                {\includegraphics[height=3.19cm]{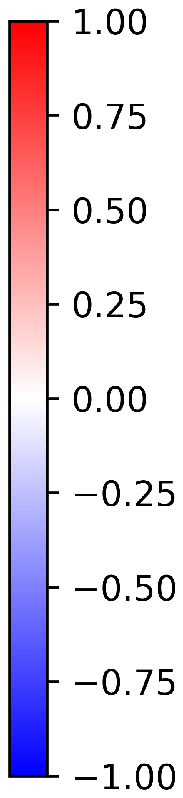}}
        \end{subfigure}%
        \caption{
                (a,b)~Sagittal slices of reference and template image;
                (c)~overlay image before registration;
                (d)~after deformable registration, the overlay image clearly highlights morphological differences; (e)~the difference image between GPU- and OMP-based registration results shows slight variations in regions with few unambiguous correspondences, such as the colon; (f) final registration result with differences highlighted in color. 
                Image courtesy of  Radboud University Medical Center, Nijmegen, Netherlands.}
        \label{2653fig:ct_colon}
\end{figure}

It is prudent to ask whether moving from double precision to single precision on the GPU introduces differences due to rounding. In fact, we observed that this can have an effect (Figure~\ref{2653fig:ct_colon}). However, it typically only occurs when there are no clear correspondences, such as in regions of the colon with different content, or when the examination table is visible in one of the two scans. In these areas, there is no strong objective function gradient in either direction during optimization, so that numerical differences have a larger impact. However, we argue that if such areas were to be registered accurately, a more elaborate model that accounts for the possible removal of  structures would have to be employed in any case.

\subsection{DIR-Lab 4DCT benchmark}
For a comparison to the state of the art, we evaluated our method on the DIR-Lab 4DCT dataset \cite{2653-08,2653-09}, consisting of ten CT scan pairs of the lung in fully-inhaled and fully-exhaled state. Resolutions are in the range of $256^2 \times \{94, \ldots, 112\}$ for the first five images and $512^2 \times \{120, \ldots, 136\}$ for the last five images. We set the deformation grid to one quarter of the image resolution.

\begin{figure}[t]
        \caption{Comparison of average \textit{landmark error} (LME) in mm and runtime based on the DIR-Lab dataset. Shown are the algorithms with smallest average LME. 
        While achieving state-of-the-art accuracy, our method is faster by orders of magnitude and provides fully deformable 3D registrations in $0.32$ seconds on average.
        }
        \label{2653fig:dirlab}
        \centering        
        \includegraphics[height=0.355\textwidth]{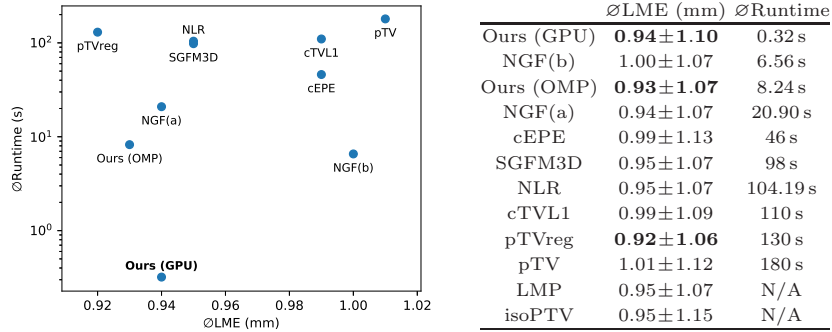}\quad\quad
        {\scriptsize
            \begin{tabular}[b]{@{}ccc@{}}
                \hline
                & $\varnothing$LME (mm) & $\varnothing$Runtime\\ 
                \hline
                Ours (GPU) & $\mathbf{0.94\!\pm\!1.10}$ & 0.32\,s\\ 
                NGF(b) & $1.00\!\pm\!1.07$ & 6.56\,s\\
                Ours (OMP) & $\mathbf{0.93\!\pm\!1.07}$ & 8.24\,s\\ 
                NGF(a) & $0.94\!\pm\!1.07$ & 20.90\,s\\ 
                cEPE & $0.99\!\pm\!1.13$ & 46\,s\\
                SGFM3D & $0.95\!\pm\!1.07$ & 98\,s\\ 
                NLR & $0.95\!\pm\!1.07$ & 104.19\,s\\ 
                cTVL1 & $0.99\!\pm\!1.09$ & 110\,s\\
                pTVreg & $\mathbf{0.92\!\pm\!1.06}$ & 130\,s\\ 
                pTV & $1.01\!\pm\!1.12$ & 180\,s\\
                LMP & $0.95\!\pm\!1.07$ & N/A\\
                isoPTV & $0.95\!\pm\!1.15$ & N/A\\
                \hline 
            \end{tabular}
            }
\end{figure}

Accuracy of the final registration was measured by the average \textit{landmark error}~(LME) over 300 expert-annotated landmarks for each dataset (Figure~\ref{2653fig:dirlab}).
Our OMP implementation scores only slightly behind the best-performing pTVreg method at an average LME of $0.92$~mm vs.~$0.93$~mm and places second-best overall in terms of accuracy.

Our GPU implementation follows closely due to the single precision computations and achieves third place overall in terms of accuracy at an LME of $0.94$~mm. Moreover, it is about one order of magnitude faster than all other methods in the benchmark for which runtimes could be obtained. Compared to the only method with better accuracy (pTVreg), it is approximately~$400$ times faster, at an average of $0.32$~seconds per full 3D registration. 


\section{Discussion}
We introduced a new method for non-linear registration using the GPU, which is highly efficient while maintaining state-of-the-art accuracy. We compared it to an optimized CPU implementation and achieved speedups up to a factor of $32.53 \pm 10.04$ at runtimes under $2$ seconds, while placing third with respect to accuracy in the DIR-Lab 4DCT benchmark. We believe that such low overall runtimes will open up new application scenarios for clinical use, such as interactive registration and  real-time organ tracking, and will further clinical adoption of fully-deformable, non-rigid registration methods.

\bibliographystyle{./bvm2019}
\bibliography{2653}

\begin{thebibliography}{1}

\bibitem{2653-01}
K\"{o}nig L, R\"{u}haak J.
\newblock A fast and accurate parallel algorithm for non-linear image
  registration using Normalized Gradient fields.
\newblock Proc IEEE Int Symp Biomed Imaging. 2014 apr; p. 580--583.

\bibitem{2653-02}
K\"{o}nig L, et~al.
\newblock A Matrix-Free Approach to Parallel and Memory-Efficient Deformable
  Image Registration.
\newblock SIAM J Sci Comput. 2018 jan;40(3):B858--B888.

\bibitem{2653-03}
Meike M.
\newblock {GPU-basierte nichtlineare Bildregistrierung} [mathesis].
\newblock Universit\"{a}t zu L\"{u}beck; 2016.

\bibitem{2653-04}
Modersitzki J.
\newblock {FAIR: Flexible Algorithms for Image Registration}.
\newblock SIAM; 2009.

\bibitem{2653-05}
Fischer B, Modersitzki J.
\newblock A unified approach to fast image registration and a new curvature
  based registration technique.
\newblock Linear Algebra Appl. 2004 mar;380:107--124.

\bibitem{2653-06}
Nocedal J.
\newblock Updating Quasi-Newton Matrices with Limited Storage.
\newblock Mathematics of Computation. 1980 sep;35(151):773--782.

\bibitem{2653-07}
Wilt N.
\newblock {The CUDA Handbook: A Comprehensive Guide to GPU Programming}.
\newblock Addison-Wesley; 2013.

\bibitem{2653-08}
Castillo R, et~al.
\newblock A framework for evaluation of deformable image registration spatial
  accuracy using large landmark point sets.
\newblock Phys Med Biol. 2009 mar;54(7):1849--1870.

\bibitem{2653-09}
Castillo E, et~al.
\newblock Four-dimensional deformable image registration using trajectory
  modeling.
\newblock Phys Med Biol. 2009 dec;55(1):305--327.

\end{thebibliography}

\end{document}